\documentclass[a4paper,fleqn]{cas-dc}

\usepackage{booktabs}
\usepackage{enumitem}
\usepackage{subcaption}
\usepackage{tabularx}

\usepackage[authoryear]{natbib}

\def\tsc#1{\csdef{#1}{\textsc{\lowercase{#1}}\xspace}}
\tsc{WGM}
\tsc{QE}
\tsc{EP}
\tsc{PMS}
\tsc{BEC}
\tsc{DE}
\begin{document}
\let\WriteBookmarks\relax
\def\floatpagepagefraction{1}
\def\textpagefraction{.001}
\shorttitle{Multi-target prediction of CVD biomarkers}
\shortauthors{Inekwe et~al.}

\title [mode = title]{A Multi-target Bayesian Transformer Framework for Predicting  Cardiovascular Disease Biomarkers during Pandemics}                      
% \tnotemark[1,2]

% \tnotetext[1]{This document is the results of the research
%    project funded by the National Science Foundation.}

% \tnotetext[2]{The second title footnote which is a longer text matter
%    to fill through the whole text width and overflow into
%    another line in the footnotes area of the first page.}

% \affiliation[label1]{organization={Deptartment of Computer Science, WPI},
%             % addressline={100 Institute Rd},
%             city={Worcester},
%             postcode={01609},
%             state={MA},
%             country={USA}}

% \affiliation[label2]{organization={Department of Genetics and Computational Biology, UMass Med},
%             % addressline={55 N Lake Ave},
%             city={Worcester},
%             postcode={01655},
%             state={MA},
%             country={USA}}

% \affiliation[label3]{organization={Department of Data Science and Analytics, SUNY Polytechnic Inst.},
%             % addressline={100 Seymour Rd},
%             city={Utica},
%             postcode={13502},
%             state={NY},
%             country={USA}}

\author[1,3]{\textcolor{black}{Trusting Inekwe}}[%
    % type, role, etc. optional
    % auid=000,
    % bioid=1,
    % prefix=Mr,
    % role=First Author,
    orcid=0000-0002-1921-4426,
]
\cormark[1]  % corresponding author mark (renders as asterisk)
\fnmark[1]   % footnote mark (renders as superscript 1)
% \ead{toinekwe@wpi.edu}
% \ead[URL]{https://www.sunypoly.edu}

% \author[label1,label3]{\textcolor{black}{Trusting Inekwe \corref{cor1}\fnref{equal}}}
% \cortext[cor1]{Corresponding author}
% % \fnmark[1]
% \fnref{equal}}}

\credit{Data curation, Methodology, Formal analysis, Conceptualization, Visualization, Writing – original draft, Writing - Review \& Editing}
% % \ead{toinekwe@wpi.edu, inekwet@sunypoly.edu}
% % \ead{inekwet@sunypoly.edu}

\author[2]{\textcolor{black}{Winnie Mkandawire}}[%
    % type, role, etc. optional
    % auid=000,
    % bioid=1,
    % prefix=Ms,
    % role=First Author,
    % orcid=0000-0001-0000-0000,
]
\fnmark[1]   % footnote mark (renders as superscript 1)
% \ead{Winnie.Mkandawire@umassmed.edu}
% \author[label2]{\textcolor{black}{Winnie Mkandawire \fnref{equal}}}
% % \ead{Winnie.Mkandawire@umassmed.edu}
\credit{Data curation, Methodology, Formal analysis, Conceptualization, Visualization, Writing – original draft, Writing - Review \& Editing}
% \fntext[equal]{\textbf{Co-first authors:} These authors contributed equally to this work.}

\author[1]{\textcolor{black}{Emmanuel Agu}}
% \ead{emmanuel@wpi.edu}
% \author[label1]{\textcolor{black}{Emmanuel Agu}}
% % \ead{emmanuel@wpi.edu}
\credit{Supervision, Writing - Review \& Editing, Resources, Conceptualization for Data curation, Methodology and analysis}

\author[2]{\textcolor{black}{Andres Colubri}}
% \ead{Andres.Colubri@umassmed.edu}
% \author[label2]{\textcolor{black}{Andres Colubri}}
% % \ead{Andres.Colubri@umassmed.edu}
\credit{Supervision, Resources, Software.}

\affiliation[1]{organization={Department of Computer Science, WPI},
    city={Worcester},
    postcode={01609},
    state={MA},
    country={USA}}

\affiliation[2]{organization={Department of Genetics and Computational Biology, UMass Chan Medical School},
    city={Worcester},
    postcode={01655},
    state={MA},
    country={USA}}

\affiliation[3]{organization={AI Exploration Center, Department of Data Science and Analytics, SUNY Polytechnic Institute.},
    city={Utica},
    postcode={13502},
    state={NY},
    country={USA}}

\cortext[cor1]{Corresponding author: toinekwe@wpi.edu, inekwet@sunypoly.edu}
\fntext[fn1]{\textbf{Co-First Authors:} These authors contributed equally to this work.}

\begin{abstract}
The COVID-19 pandemic disrupted healthcare systems worldwide, disproportionately impacting individuals with chronic conditions such as cardiovascular disease (CVD). These disruptions—through delayed care and behavioral changes—affected key CVD biomarkers, including LDL cholesterol (LDL-C), HbA1c, BMI, and systolic blood pressure (SysBP). Accurate modeling of these changes is crucial for predicting disease progression and guiding preventive care. However, prior work has not addressed multi-target prediction of CVD biomarker from Electronic Health Records (EHRs) using machine learning (ML), while jointly capturing biomarker interdependencies, temporal patterns, and predictive uncertainty.
In this paper, we propose MBT-CB, a Multi-target Bayesian Transformer (MBT) with pre-trained BERT-based transformer framework to jointly predict LDL-C, HbA1c, BMI and SysBP CVD biomarkers from EHR data. The model leverages Bayesian Variational Inference to estimate uncertainties, embeddings to capture temporal relationships and a DeepMTR model to capture biomarker inter- relationships. We evaluate MBT-CT on retrospective EHR data from 3,390 CVD patient records (304 unique patients) in 
Central Massachusetts 
% [*blinded for review*] 
during the Covid-19 pandemic. 
MBT-CB outperformed a comprehensive set of baselines including other BERT-based ML models, achieving an MAE of 0.00887, RMSE of 0.0135 and MSE of 0.00027, while effectively capturing data and model uncertainty, patient biomarker inter-relationships, and temporal dynamics via its attention and embedding mechanisms. MBT-CB's superior performance highlights its potential to improve CVD biomarker prediction and support clinical decision-making during pandemics. 
\end{abstract}

% \includegraphics{cas-grabs.pdf}

% \begin{highlights}
% \item Research highlights item 1
% \item Research highlights item 2
% \item Research highlights item 3
% \end{highlights}


\maketitle

\section{Introduction}
\textbf{Motivation: }The COVID-19 pandemic caused unprecedented disruptions to healthcare systems worldwide, disproportionately affecting individuals with chronic illnesses such as cardiovascular diseases (CVDs) \citep{verity2020estimates}. These disruptions included reduced access to medical services \citep{tsai_association_2022}, delays in routine preventive care \citep{bilgin2021missing}, and pandemic-induced lifestyle changes \citep{mattioli_covid-19_2020}. Such interruptions had a profound impact on patient outcomes, particularly affecting critical biomarkers such as LDL cholesterol (LDL-C), Glycated hemoglobin (HbA1c), BMI and Systollic Blood Pressure (SysBP)~\citep{aparisi_low-density_2021}. The ability to model and predict such  pandemic-related biomarker changes can enhance preventive care strategies and early detection of diseases.

\begin{figure*}[h]
    \centering
    \begin{subfigure}[b]{0.46\linewidth}
        \includegraphics[width=\linewidth]{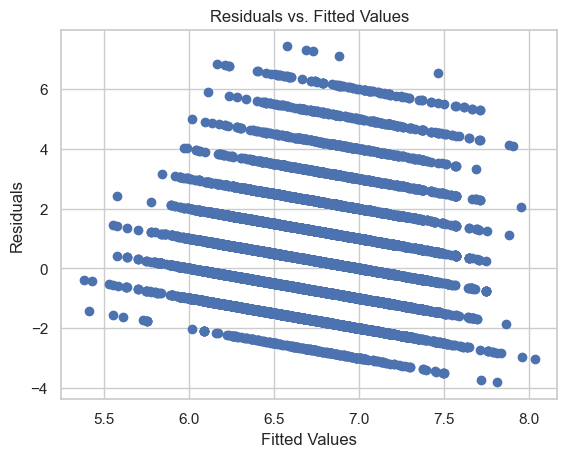}
        \caption{HbA1c showing non-constant residual variance}
        \label{fig:Heteroscedasticity_HbA1c}
    \end{subfigure}
    \hspace{0.04\linewidth}
    \begin{subfigure}[b]{0.46\linewidth}
        \includegraphics[width=\linewidth]{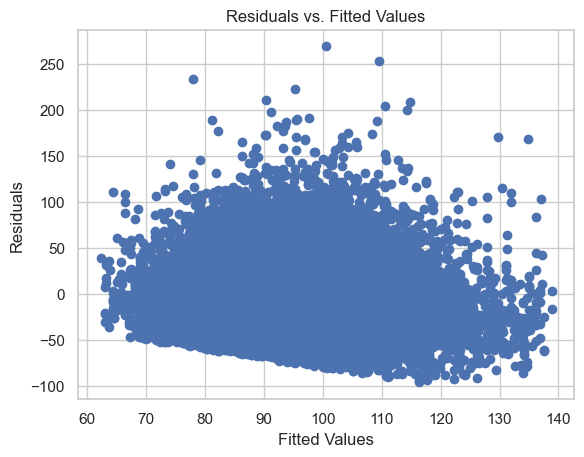}
        \caption{LDL-C showing non-constant residual variance}
        \label{fig:Heteroscedasticity_LDLChol}
    \end{subfigure}
    \caption{Heteroscedasticity in biomarkers: (a) HbA1c, (b) LDL-C.}
    \label{fig:Heteroscedasticity_combined}
\end{figure*}

\textbf{Challenges:} \textit{EHR data has temporal structure and can be recorded at irregular intervals}, introducing uncertainty~\citep{xiao2018opportunities}, which complicates predictive modeling. Irregular visits and missing or erroneous entries lead to \textit{aleatoric} (data noise) and \textit{epistemic} (model-related) uncertainty. Estimating these is challenging in deep learning and requires approaches such as \textit{Bayesian inference}~\citep{liu2020simple}, which is rarely integrated into standard attention models~\citep{kostenok2023uncertainty}. Additionally, \textit{CVD biomarkers—such as HbA1c and LDL-C—often exhibit heteroscedasticity}, with non-constant variance (Figures~\ref{fig:Heteroscedasticity_HbA1c},~\ref{fig:Heteroscedasticity_LDLChol}). Addressing this requires models that learn across noisy outputs. \textit{CVD biomarkers are also interdependent}~\citep{wu1998relationship}, reflecting shared physiological and clinical factors. Thus, models that jointly model temporal dynamics, output dependencies, and uncertainty are critical for reliable prediction during crisis periods like pandemics.

\textbf{Our approach:} We propose MBT-CB NN, a novel transformer-based architecture for modeling temporal relationships in patient EHR data, multi-target biomarker prediction, and uncertainty estimation. Building on the ClinicalBERT model
% ~\citep{wang2023optimized, 
\citep{liu2025generalist} via fine-tuning, MBT-CB integrates Bayesian Variational Inference (BVI) to capture aleatoric and epistemic uncertainty~\citep{lauret2008bayesian}, a Deep Multi-Target Regression (DeepMTR) layer~\citep{reyes2019performing} to learn shared and target-specific representations, and positional embeddings~\citep{vaswani2017attention}, for temporal encoding. Trained on 3,390 EHRs (304 unique patients) from 
Central Massachusetts 
% [*blinded for review*] 
between Jan 2019–June 2021, MBT-CB is tailored for CVD biomarker prediction during the pandemic.

\textbf{Novelty of Work: } While transformers have been used for CVD detection \citep{antikainen2023transformers},  multi-target prediction \citep{poulain2021transformer}, and uncertainty-aware COVID-19 models \citep{chen2023covid}, no prior work combines BVI, multi-target prediction, and temporal attention in a unified transformer framework for longitudinal CVD biomarker prediction. Our model delivers robust, confidence/uncertainty-aware predictions suitable for high-stakes clinical use.
\noindent\textbf{Our Contributions:}
\begin{enumerate}[leftmargin=5pt]
    \item \textit{Bayesian Multi-Target Transformer:} We propose MBT-CB, a novel transformer architecture with BVI that jointly predicts LDL-C, HbA1c, BMI, and SysBP from EHR data while modeling uncertainty/confidence, temporal dynamics, and biomarker interdependencies.

    \item \textit{Systematic Rigorous evaluation:} We benchmark MBT-CB against other models. MBT-CB outperforms traditional ML, DL baselines, and pretrained transformers, achieving MAE 0.00887, RMSE 0.0135, and MSE 0.00027, showing high accuracy and generalizability under pandemic-related variability.

    \item \textit{Explainability via attention and uncertainty visualization:} MBT-CB reveals dependencies such as HbA1c–BMI, highlights  aleatoric and epistemic uncertainty, supporting transparent, risk-aware predictions.
\end{enumerate}

\section{Related Work}
\subsection{Transformers for COVID-19 and CVD prediction}

\noindent\textbf{CVD detection using Transformers on EHR \& Image datasets: }Antikainen et al., \citep{antikainen2023transformers} compared BERT and XLNet for mortality prediction in 23,000 cardiac patients using EHR time series data. XLNet slightly outperformed BERT, with an AUC of 76.0\% vs. 75.5\%, and showed a 9.8\% higher recall, indicating better identification of at-risk patients.
Similarly, Kilimci et al \citep{kilimci2023heart} explored deep vision transformers (Google-ViT, Microsoft-Beit, and Swin-Tiny) for heart disease detection from ECG images, achieving high accuracies Swin-Tiny (95.9\%), Microsoft-Beit (95.5\%) and Google-ViT (94.3\%).
\par\noindent\textbf{Transformers and Multi-target Prediction:} Poulain et al. \citep{poulain2021transformer}, enhanced BERT with a Deep Multi-Target Regression (DeepMTR) module to predict 11 modifiable CVD risk factors. The model significantly improved RMSE (by 0.053) for targets with over 80\% missing data and reduced MAE by an average of 12.6\% across all targets compared to baselines.

\noindent\textbf{COVID detection using transformers and Bayesian NN):} Chen et al., \citep{chen2023covid} applied a hybrid transformer model (CBAM, ViT, Swin Transformer) to chest X-rays for COVID-19 detection, adding a Bayesian NN layer to capture weight uncertainty. Unlike their approach that adds Bayesian methods externally, our study embeds Bayesian inference directly into the attention mechanism to quantify uncertainty in CVD prediction from EHR data.

\noindent\textbf{Transformers and Bayesian approach: } Sankararaman \citep{sankararaman2022bayesformer} investigated a Bayesian transformer approach by treating learnable parameters as random distributions. In contrast, our approach selectively applies Bayesian methods to attention weights, treating them as random variables while keeping other parameters constant.

\textbf{Research gap addressed} 
Although Bayesian and multi-target transformers have been applied to clinical prediction, no prior work has unified multi-target learning, temporal modeling, and uncertainty estimation within a single framework to predict the impact of COVID-19 on key CVD biomarkers (HbA1c, LDL, BMI, and SysBP). Addressing these gaps is essential for reliable deployment in high-stakes healthcare settings affected by pandemic-related care disruptions.

\section{Methodology}

\begin{figure*}[htbp]
\centering
\includegraphics[width=\linewidth]{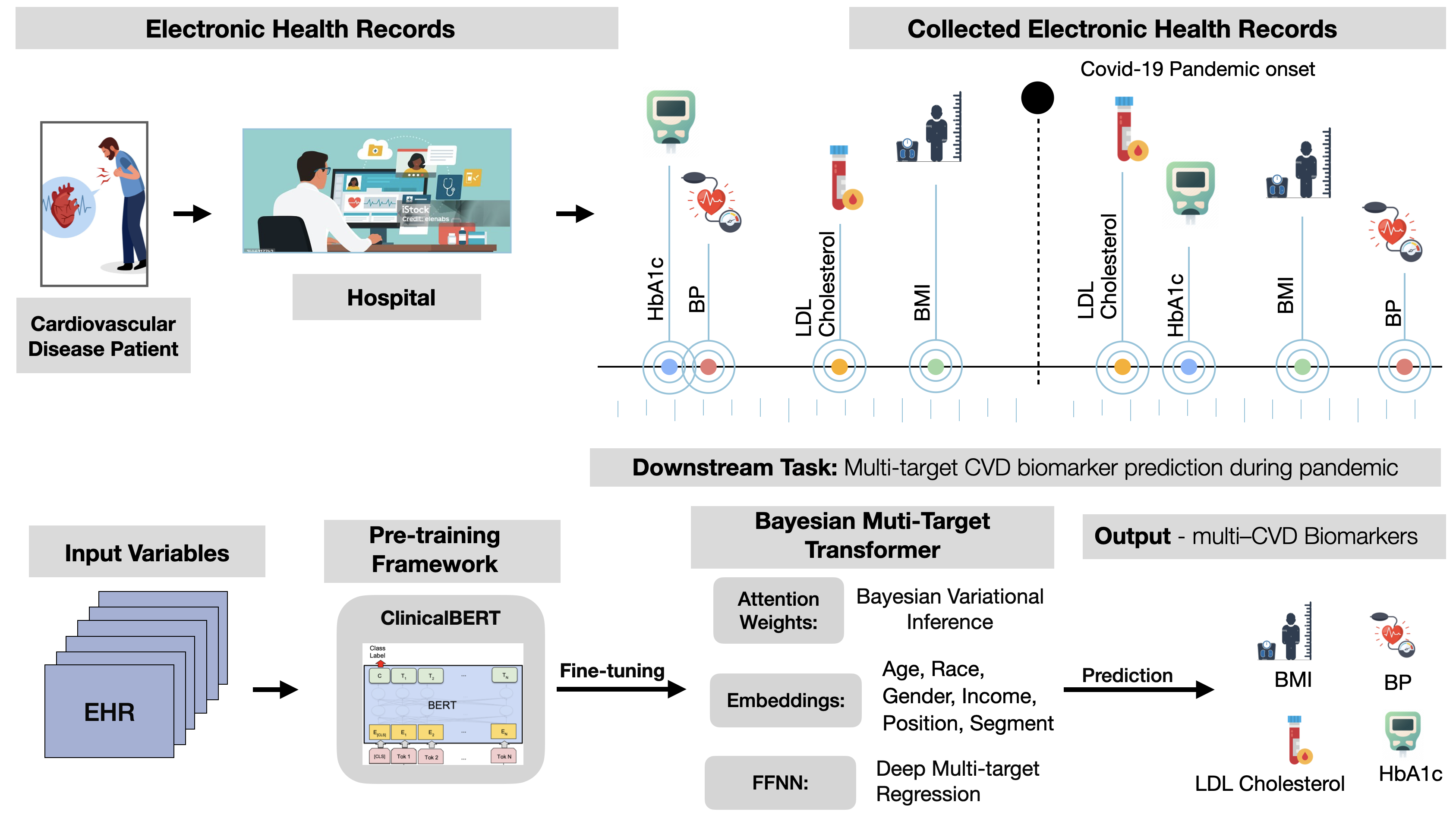}
\caption{Overview of Methodology}
\label{fig:overview_of_meth}
\end{figure*}

\subsection{Overview of Methodology}
Figure~\ref{fig:overview_of_meth} outlines MBT-CB, our framework for predicting CVD biomarker trajectories during the COVID-19 pandemic. We utilized EHR data from 304 CVD patients treated at 
UMass Chan and Memorial Hospital, 
% [*blinded for review*] 
spanning pre-pandemic (Jan 2018–Dec 2019) and pandemic (June 2020–June 2021) periods. The dataset included demographics (age, gender, income, race) and biomarkers (HbA1c, LDL, BMI, SysBP). To predict biomarker values at the first pandemic-era visit, biomarker data were encoded using ClinicalBERT representations. The model was then fine-tuned on this cohort to capture population-specific patterns and temporal shifts in biomarker dynamics. Variational inference was applied to attention weights, with embeddings and outputs processed by the DeepMTR FFNN model.
\par
\begin{figure*}[htbp]
    \centering
    \includegraphics[width= 0.5\linewidth]{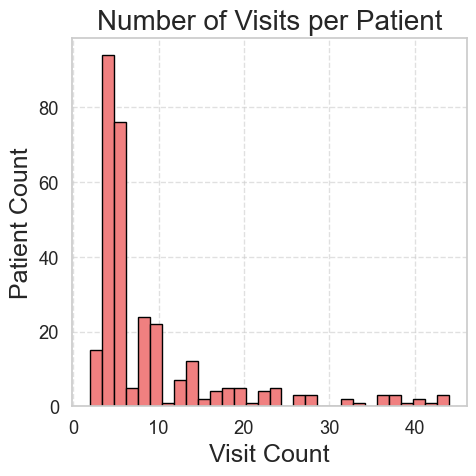}
    \caption{Visit frequency of patients} 
    \label{fig:no_of_visit}
\end{figure*}

\textbf{Data Pre-processing}
First, patients with CVD were identified using ICD-10 codes (I63, I67, I48, I73, I25, I65, I50). Records with nulls, outliers, implausible values, or no hospital visits during the COVID-19 pandemic were removed, reducing the dataset from 80,917 to 3,390 records across 304 unique patients. Visit frequency distribution is shown in Figure~\ref{fig:no_of_visit}. 

\par\textbf{Feature Extraction and Engineering}
\label{sec:Feature_Ext}
HbA1c, LDL-C, BMI, and SysBP were selected as target variables. Socioeconomic status—estimated via median household income by zip code~\citep{us_zip_codes}—was included due to its impact on health outcomes~\citep{hawkins2020socio}. Categorical features (gender, race, income) were one-hot encoded, numerical features were normalized, and skewed distributions log-transformed. The final dataset (Table~\ref{table:InOutvariables}) included both categorical and numerical features.

\begin{table*}[htbp]
\centering
\caption{Dataset Variables with Descriptions and Distributions}
\label{table:InOutvariables}
\setlength{\tabcolsep}{3pt}
\renewcommand{\arraystretch}{0.90}
{\footnotesize
\begin{tabularx}{\textwidth}{p{3.7cm} p{5.3cm} p{5.3cm}}
\toprule
\textbf{Feature} & \textbf{Description} & \textbf{Range / Distribution}\\
\midrule
\makecell[l]{Glycated \\ Hemoglobin (HbA1c)}
 & \makecell[l]{Average blood sugar level\\ for the past 2-3 months }& $4\% < \text{HbA1c} \leq 14\%$ \\
\midrule
\makecell[l]{LDL \\ Cholesterol} & \makecell[l]{A type of lipoprotein that \\carries cholesterol in the blood} & $20\,\text{mg/dL} < \text{LDL} \leq 370\,\text{mg/dL}$ \\
\midrule
\makecell[l]{Blood Pressure \\(BP)} & \makecell[l]{Force of blood against artery \\walls (systolic and diastolic)} & Systolic: $84 \leq \text{BP} \leq 196$ \\
\midrule
\makecell[l]{Body Mass Index \\(BMI)} & \makecell[l]{Measure of weight in relation \\to height} & $15 \leq \text{BMI} \leq 100$ \\
\midrule
\makecell[l]{Socioeconomic \\Status (Income)} & \makecell[l]{Position in social and economic \\hierarchy} & \makecell[l]{38.2\% upper middle class, \\ 36.8\% middle class, 24.3\% \\lower middle class, 0.7\% \\upper class} \\
\midrule
Age & Age distribution in the dataset & $45 \leq \text{Age} \leq 96$ \\
\midrule
Race & Race distribution in the dataset & \makecell[l]{88.2\% White, 7.2\% Black/\\African American, 4.6\% Asian }\\
\midrule
Gender & \makecell[l]{Biological sex recorded at time \\of visit} & 61.5\% Male, 38.5\% Female \\
\bottomrule
\end{tabularx}
}
\end{table*}

The dataset comprised four biomarkers: HbA1c (4\% to 14\%), LDL-C (20–370 mg/dL), systolic blood pressure(SysBP) (84–196 mmHg), and BMI (15–100). Demographics: income level (38.2\% upper-middle, 36.8\% middle, 24.3\% lower-middle, 0.7\% upper), age (45–96 years), race (88.2\% White, 7.2\% Black, 4.6\% Asian), and gender (61.5\% male, 38.5\% female). These variables capture both physiological and social determinants relevant to cardiovascular health.
\noindent Data was split 60/20/20 train/validation/test at patient level, preventing leakage and preserving longitudinal visit structure.

% \textbf{MBT-CB Implementation for CVD Prediction}
% details of the transformer model are in Appendix~\ref{sec:van_transformer}.
Details on the transformer architecture, Bayesian Methods in DL and Variational Inference for Attention Weights are provided in \ref{sec:van_transformer},\ref{sec:bayesianNN} and \ref{sec:var_inference} respectively. Below, we describe our MBT-CB Implementation for CVD Prediction. 
\begin{figure*}[htbp]
\centering
\includegraphics[width=\linewidth]{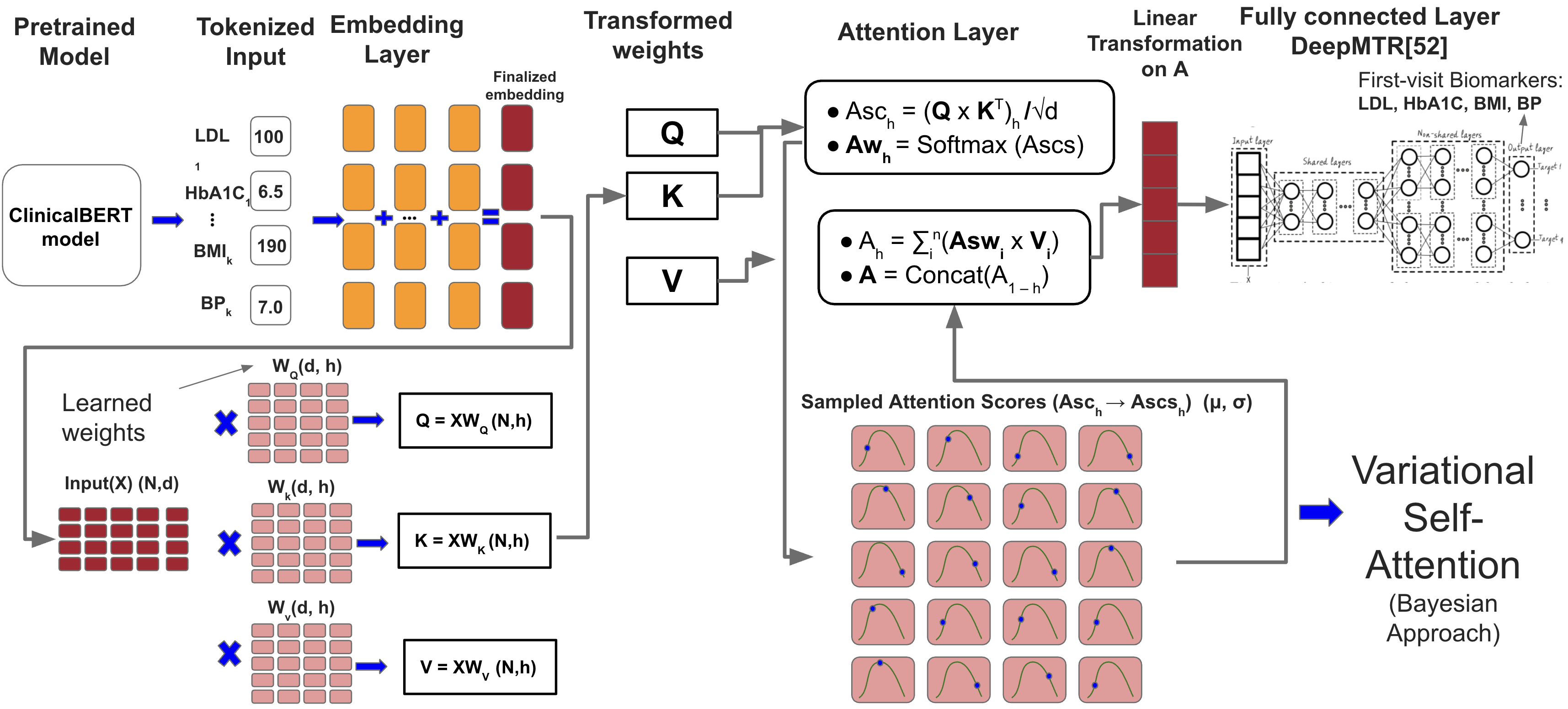}
\caption{Our proposed MBT-CB framework based on a Transformer with Variational Self Attention. Patient's EHR biomarker values for $k$ visits (1 to $k$) are passed as input variables to the MBT-CB model. Prediction is on the $1^{st}$ visit, where $k < n$. DeepMTR image from \citep{reyes2019performing}}
\label{fig:transformer_Var_inf_Att}
\end{figure*}
\subsubsection{\textbf{Input sequence and patient representation}}
\label{sec:input_representation}

\begin{figure*}[htbp]
\centering
\includegraphics[width =\textwidth]{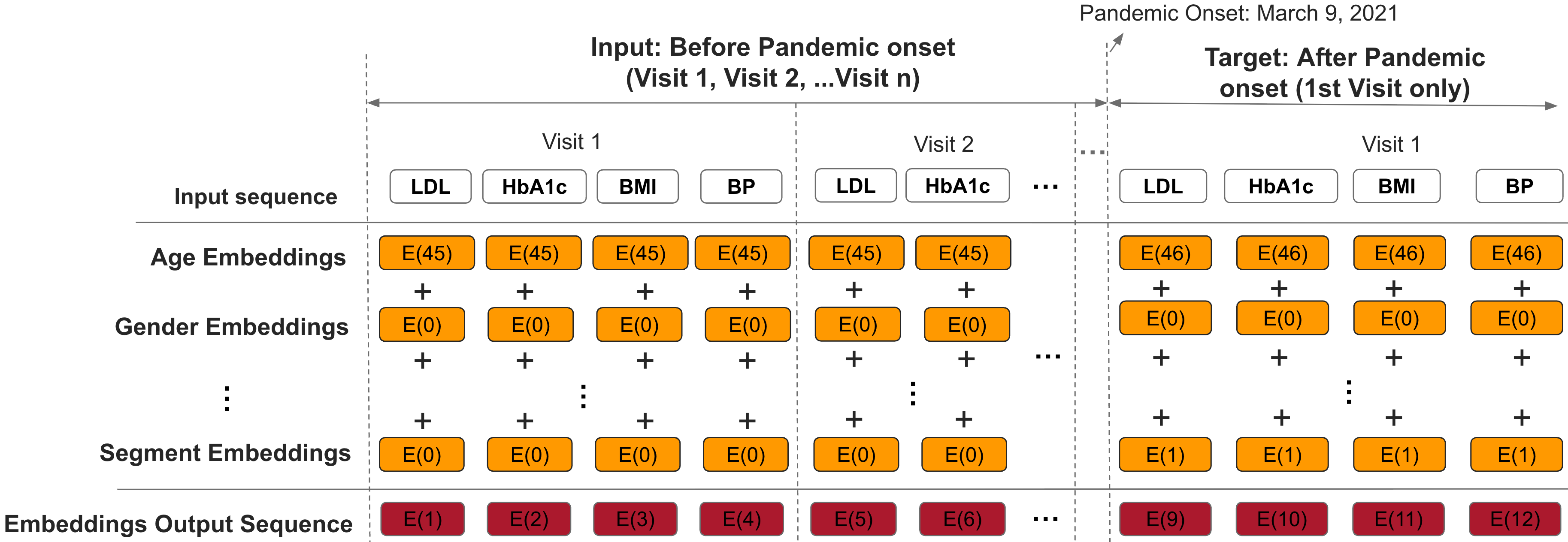}
\caption{Modification of a patient's EHR record. Each row represents a set of chronological biomarker values from an individual clinical visit, suitable for transformer-based modeling.}
\label{fig:dataset_modification}
\end{figure*}

As shown in Figure \ref{fig:dataset_modification}, each patient's EHR is structured as a chronological sequence of clinical visits, forming a time series of four key CVD biomarkers: SysBP, BMI, HbA1c, and LDL-C. For patient $p$, the visit sequence is denoted as $\mathbf{V}_p = \{ v_p^1, v_p^2, \dots, v_p^{n_p} \}$, where $n_p$ is the number of visits. Each visit $v_p^i$ is transformed into a structured sentence: $T_p^i = \text{``sys: } x_{\text{sys}}^i \text{; bmi: } x_{\text{bmi}}^i \text{; hba1c: } \newline x_{\text{HbA1c}}^i \text{; chol: } x_{\text{chol}}^i \text{''}$, where each $x^i$ is a normalized scalar biomarker value. These text sequences are input into a pre-trained transformer for contextual representation learning.
\subsubsection{\textbf{Transfer Learning}} 
\label{transfer_learning}
% \hfill \break
\textbf{Pre-trainng: }To leverage existing domain-specific knowledge, a transfer learning strategy using ClinicalBERT~\citep{wang2023optimized, liu2025generalist} is adopted.
% , a pretrained language model based on BERT and fine-tuned on a large EHR corpus. 
ClinicalBERT is a language model pre-trained on over 3million structured EHR records using the masked language modeling objective on over 1.2billion words of clinical text.

\noindent\textbf{Fine-Tuning: }Each input sentence $T_p^i$ is tokenized using the ClinicalBERT's tokenizer and passed through the encoder to produce a contextualized embedding from the [CLS] token:
$\mathbf{h}_p^i = \text{Model}_{\text{pre}}(T_p^i)_{\texttt{[CLS]}} \in \mathbb{R}^d$
This contextual embedding is then projected into a task-specific representation using a linear transformation:
$\mathbf{z}_p^i = \mathbf{W}_{\text{proj}} \cdot \mathbf{h}_p^i + \mathbf{b}_{\text{proj}}$. To enhance these representations, we append several auxiliary embeddings:

\begin{itemize}[leftmargin=5pt]
    \item \textbf{Positional embeddings} $\text{pos}_p = \{0, 1, \dots, n_p - 1\}$ to capture visit order,
    \item \textbf{Segment embeddings} $\text{seg}_p = \{s_p^1, \dots, s_p^{n_p}\}$ where $s_p^i \in \{0, 1\}$ indicates whether the visit occurred before or after the onset of COVID-19,
    \item \textbf{Demographic identifiers}: gender $g_p$, race $r_p$, and income class $i_p$, encoded as categorical indices.
\end{itemize}

To enable the transformer model to jointly learn from temporal visit patterns, biomarker trajectories, and relevant demographic context, each patient $p$'s complete input is represented as 
    $\text{Input}_p = \left( \{T_p^i\}_{i=1}^{n_p},\ \text{pos}_p,\ \text{seg}_p,\ g_p,\ r_p,\ i_p \right)$.
\subsubsection{\textbf{Variational Self-Attention Mechanism}}
To capture epistemic uncertainty, the fixed projection weights are replaced with a Gaussian distribution parameterized by a learnable mean $\mu$ and log standard deviation 
$\sigma$. Weights are sampled during each forward pass: $W = \mu + \exp(\log \sigma) \cdot \varepsilon, \quad \varepsilon \sim \mathcal{N}(0, I)$, Keys are transformed as $k' = k W$, and attention scores computed via: $S = \frac{q \cdot {k'}^\top}{\sqrt{d}}$. More details in Appendix Sections \ref{sec:bayesianNN} and \ref{sec:var_inference} respectively
\subsubsection{\textbf{Bayesian Prior and Objective Function}}
% \subsubsection{Objective Function}
A standard normal prior is placed on $W$, regularized via KL divergence. The training loss is $\mathcal{L}_{\text{total}} = \text{MSE}(y, \hat{y}) + \lambda \cdot \text{KL}(q(W) \parallel p(W))$ where
$\text{KL} = \sum \left[ -\log \sigma + \frac{1}{2} \left( \sigma^2 + \mu^2 - 1 \right) \right]$.
\subsubsection{\textbf{DeepMTR}}
To capture both shared and biomarker-specific patterns, the attention output feeds into DeepMTR, which includes shared layers and target-specific heads for each biomarker. 
\textbf{The MBT-CB Model Pipeline}
restructures patient visit histories into biomarker sentences, tokenized to generate contextual embeddings using ClinicalBERT. Positional, segment, and demographic embeddings are added to enrich the representation, and Variational Self-Attention applied to model epistemic uncertainty. The resultant representations are input to a DeepMTR head for multi-biomarker prediction, with uncertainty quantified via multiple stochastic forward passes.
\section{Evaluation}
\label{sec:Evaluation Metrics}

\textbf{Evaluation metrics: }used to assess MBT-CB's performance on a test set included MAE, MSE and RMSE across multiple CVD biomarkers. The mathematical expressions of these metrics are %in Appendix~\ref{eval_metrics}.
shown below:
% \subsection{Evaluation Metrics}
\label{eval_metrics}
    \begin{equation}
        \text{MAE} = \frac{1}{n} \sum_{i=1}^{n} |Y_i - \hat{Y}_i|
    \end{equation}

    \begin{equation}
        \text{MSE} = \frac{1}{n} \sum_{i=1}^{n} (Y_i - \hat{Y}_i)^2
    \end{equation}

    \begin{equation}
        \text{RMSE} = \sqrt{\frac{1}{n} \sum_{i=1}^{n} (Y_i - \hat{Y}_i)^2}
    \end{equation}
\textbf{Baseline Models} were selected from the following categories: Bio\_ClinicalBERT, MedBERT, BERT BERT-based models (default, and also integrated with BVI and DeepMTR), a FFNN and a Linear regression. All models were configured for multi-target prediction and are described briefly. 

\begin{enumerate}[leftmargin=5pt]
\item \textit{Bio\_ClinicalBERT:} derived from BioBERT~\citep{lee2020biobert}, was trained on clinical notes from MIMIC III EHR dataset~\citep{johnson2016mimic} and has approximately  880 million words. 
 \item \textit{MedBERT~\citep{9980157}: } derived from Bio\_ClinicalBERT, was pre-trained on over 57.46 million tokens from multiple medical datasets including the N2C2, BioNLP project4 and Wikipedia corpora.
    
    \item \textit{BERT~\citep{devlin2019bert}: }  is a language model pre-trained on a large corpus of 11,038 unpublished books and Wikipedia in English. BERT-base-uncased,  trained on 110 million words, was utilized. 

    \item \textit{Traditional ML models: } including a multi-target FFNN and Linear Regression models.

\end{enumerate}

In the results section, performance is reported across all four pre-trained model variants. The best-performing variant is selected for final evaluation and uncertainty quantification.
\textit{An ablation study} was conducted study to evaluate the contributions of the DeepMTR and BVI modules to model performance.

\section{Results}
\subsection{MBT-CB prediction performance and comparisons}
As shown in Table \ref{tab:mae_rmse_mse_models_multi-target},
% mae_rmse_models_multi-target} and
% \ref{tab:mae_models_multi-target}, \ref{tab:rmse_models_multi-target} and 
% \ref{tab:mse_models_multi-target}, 
the complete (proposed) MBT-CB model outperformed all baselines in multi-target prediction of CVD biomarkers during the pandemic with a lowest mean value for MAE (0.00887), RMSE (0.0135) and MSE (0.00027). 

\textbf{1. Comparisons to other integrated pre-trained transformer models (bayesian and DeepMTR): } MBT-CB outperformed MedBERT, Bio\_ClinicalBERT and BERT integrated pretrained transformer, achieving an MAE of 0.0051, 0.0071, 0.0173, and 0.0059 for SysBp, BMI, HbA1c, and LDL-C respectively. Corresponding RMSE values were 0.0065, 0.0094, 0.0298, and 0.0082, while MSE values remained consistently low across all targets (SysBp: 0.000042, BMI: 0.000088, HbA1c: 0.000890, LDL-C: 0.000067), indicating minimal error variance and strong prediction stability. Bio\_ClinicalBERT was the next best performing model with a mean MAE of 0.0138, MSE(0.00044) and RMSE (0.0185). MedBERT followed with a mean MAE of 0.1450, MSE(0.00048) and RMSE (0.0199). Lastly, with a mean MAE of 0.0169, MSE (0.00055) and RMSE (0.0217), BERT exhibited the lowest performance in this group. These results reveal that incorporating Bayesian inference and the DeepMTR architecture into pre-trained clinical language models enhances accuracy, particularly in predicting multi-target biomarker trajectories.

\textbf{2. Comparisons to non-integrated transformer-based baselines:}  While the non-integrated pretrained models performed reasonably well across all biomarkers, they were still outperformed by our proposed MBT-CB model. This suggests that there is an added benefit in using a fully end-to-end Bayesian Transformer architecture with integrated multi-target regression  capabilities to enhance predictive accuracy and generalization across multiple CVD biomarkers.

\textbf{3. Comparisons with non-transformer models:} such as multi-target linear regression and multi-target FFNN revealed that they were unable to generalize, with MAE values exceeding 0.5 for most targets and MSE values as high as 1.02. This highlights the limitations of non-transformer approaches in handling complex, longitudinal EHR data with interrelated targets.

\textbf{4. Ablative analyses:} 
The full MBT-CB model outperformed all ablation variants across all biomarkers. Removing BVI led to higher MAEs, except for LDL-C, which saw slight improvements (MAE: 0.0039, RMSE: 0.00576, MSE: 0.000033). Still, the complete model achieved the best overall performance, highlighting the value of Bayesian attention. Excluding the DeepMTR head resulted in a substantial performance drop (mean MAE: 0.0245, RMSE: 0.033, MSE: 0.0011), confirming its critical role in modeling biomarker interdependencies.

\begin{figure*}[htbp]
    \centering
    \includegraphics[width=0.8 \linewidth]{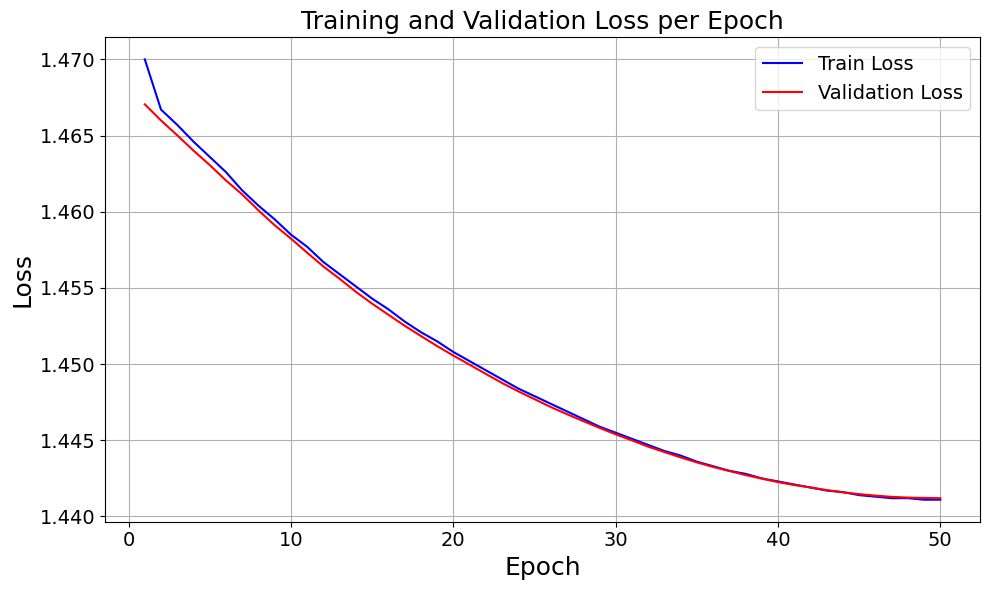}
    \caption{Loss curve of MBT-CB model}
    \label{fig:loss_curve}
\end{figure*}

\textbf{5. Training dynamics and attention visualization:} As shown in Figure~\ref{fig:loss_curve}, the MBT-CB model’s training and validation losses steadily decreased over 50 epochs without overfitting. 

\begin{figure*}[htbp]
\centering
\includegraphics[width= 0.3\textwidth]{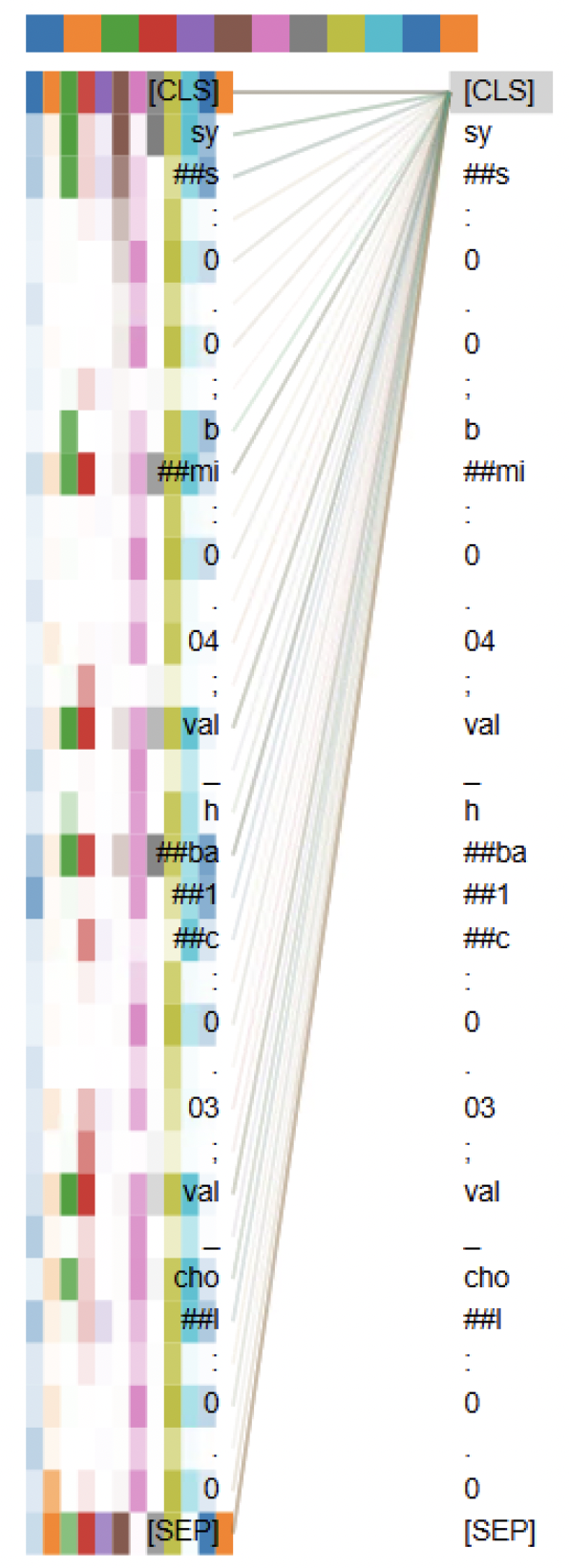}
\caption{Attention relationship visualization}
\label{fig:attention_viz}
\end{figure*}

Figure~\ref{fig:attention_viz} visualizes the attention pattern of MBT-CB's Bayesian attention mechanism. The attention scores display token-level dependencies across visits and biomarkers, suggesting that MBT-CB learns nuanced patterns, such as the relationship between HbA1c and BMI trajectories, structured, temporal input.

\textbf{Summary of findings:} The MBT-CB architecture— with BVI self-attention and a multi-target regression head— accurately predicted multiple CVD biomarker trajectories and substantially outperformed all baselines, while supporting interpretation via attention patterns. %Details on hardware and computational resources used are in Appendix \ref{sec:comp_compl}

\subsection{Computational Complexity and Feasibility}
\label{sec:comp_compl}
All experiments were run on Amazon WorkSpaces (EC2 m7i-flex.2xlarge) with Microsoft Windows Server 2019, 8 vCPUs (Intel Xeon Platinum 8375C @ 2.4 GHz), and 32 GB RAM, without GPU acceleration. MBT-CB trained for 50 epochs in approximately 6 hours, and inference per patient sequence required ~150–200 ms, indicating that the model is computationally efficient even on CPU-only infrastructure. For deployment in low-resource settings, model compression techniques (e.g., distillation and quantization) can further reduce runtime and resource usage, making MBT-CB practical for real-world clinical environments.

\begin{table*}[ht]
    \centering
    \caption{MAE, RMSE, and MSE Performance of Various Models for Multi-target Biomarker Prediction (all values in decimal form, normalized to exponential level $1\mathrm{e}{-2}$)}
    \label{tab:mae_rmse_mse_models_multi-target}
    \resizebox{\textwidth}{!}{%
    \renewcommand{\arraystretch}{1.0}
    \setlength{\tabcolsep}{3pt}
    \begin{tabular}{lccccccccccccccc}
        \toprule
        \textbf{Model} 
        & \multicolumn{5}{c}{\textbf{MAE}} 
        & \multicolumn{5}{c}{\textbf{RMSE}} 
        & \multicolumn{5}{c}{\textbf{MSE}} \\
        \cmidrule(lr){2-6} \cmidrule(lr){7-11} \cmidrule(lr){12-16}
        & SysBp & BMI & HbA1c & LDL & Mean 
        & SysBp & BMI & HbA1c & LDL & Mean 
        & SysBp & BMI & HbA1c & LDL & Mean \\
        \midrule
        \multicolumn{16}{l}{\textbf{Multi-target}} \\
        FFNN & 73.8 & 30.3 & 51.8 & 66.0 & 55.4 
             & 101.0 & 41.8 & 70.3 & 83.0 & 74.1 
             & 103.0 & 17.5 & 49.4 & 68.9 & 59.6 \\
        Linear Regression & 59.9 & 30.2 & 55.7 & 73.8 & 55.0 
                          & 81.5 & 42.2 & 81.6 & 89.9 & 73.8 
                          & 66.4 & 17.8 & 66.6 & 80.8 & 58.0 \\
        \midrule
        \multicolumn{16}{l}{\textbf{Standard \& Multi-target FFNN}} \\
        MedBERT & 0.537 & 0.977 & 3.620 & 0.653 & 1.450 
                & 0.725 & 1.280 & 4.940 & 0.876 & 1.960 
                & 0.0053 & 0.0163 & 0.2450 & 0.0077 & 0.0685 \\
        BERT & 0.583 & 1.030 & 3.090 & 0.623 & 1.380 
             & 0.894 & 1.430 & 4.620 & 1.090 & 2.060 
             & 0.0080 & 0.0204 & 0.2130 & 0.0118 & 0.0634 \\
        Bio\_ClinicalBERT & 0.630 & 1.290 & 3.140 & 0.477 & 1.380 
                          & 1.120 & 1.730 & 4.630 & 0.778 & 2.060 
                          & 0.0125 & 0.0301 & 0.2140 & 0.0060 & 0.0656 \\
        \midrule
        \multicolumn{16}{l}{\textbf{Bayesian \& DeepMTR}} \\
        MedBERT & 0.906 & 1.330 & 2.660 & 0.918 & 1.450 
                & 1.230 & 1.740 & 3.630 & 1.170 & 1.990 
                & 0.0151 & 0.0301 & 0.1320 & 0.0137 & 0.0477 \\
        BERT & 1.140 & 1.510 & 2.840 & 1.060 & 1.690 
             & 1.410 & 1.920 & 3.800 & 1.350 & 2.170 
             & 0.0200 & 0.0367 & 0.1450 & 0.0182 & 0.0549 \\
        Bio\_ClinicalBERT & 1.020 & 1.160 & 2.540 & 0.792 & 1.380 
                          & 1.350 & 1.510 & 3.520 & 1.030 & 1.850 
                          & 0.0181 & 0.0229 & 0.1240 & 0.0105 & 0.0439 \\
        \midrule
        \multicolumn{16}{l}{\textbf{Bayesian + DeepMTR (Proposed)}} \\
        ClinicalBERT & \textbf{0.511} & \textbf{0.714} & \textbf{1.730} & 0.592 & \textbf{0.887} 
                          & \textbf{0.652} & \textbf{0.936} & \textbf{2.980} & 0.818 & \textbf{1.350} 
                          & \textbf{0.0042} & \textbf{0.0088} & \textbf{0.0890} & 0.0067 & \textbf{0.0272} \\
        \midrule
        \multicolumn{16}{l}{\textbf{Ablation Study}} \\
        w/o Bayesian & 0.580 & 1.000 & 3.170 & \textbf{0.390} & 1.280 
                     & 0.769 & 1.330 & 4.590 & \textbf{0.576} & 1.810 
                     & 0.0059 & 0.0176 & 0.2100 & \textbf{0.0033} & 0.0593 \\
        w/o DeepMTR & 2.000 & 2.290 & 3.470 & 2.050 & 2.450 
                    & 2.610 & 2.970 & 4.490 & 2.910 & 3.300 
                    & 0.0682 & 0.0885 & 0.2020 & 0.0844 & 0.1110 \\
        \bottomrule
    \end{tabular}
    }
    {\footnotesize \textit{Note}: These normalized errors can be converted to raw clinical units using the biomarker ranges in Section \ref{sec:Feature_Ext} and Table \ref{table:InOutvariables}. They all correspond to values within clinically meaningful thresholds.}
\end{table*}

\subsection{Interpreting Uncertainty in Biomarker Predictions}
\begin{figure*}[htbp]
\centering
\includegraphics[width=0.97\textwidth]{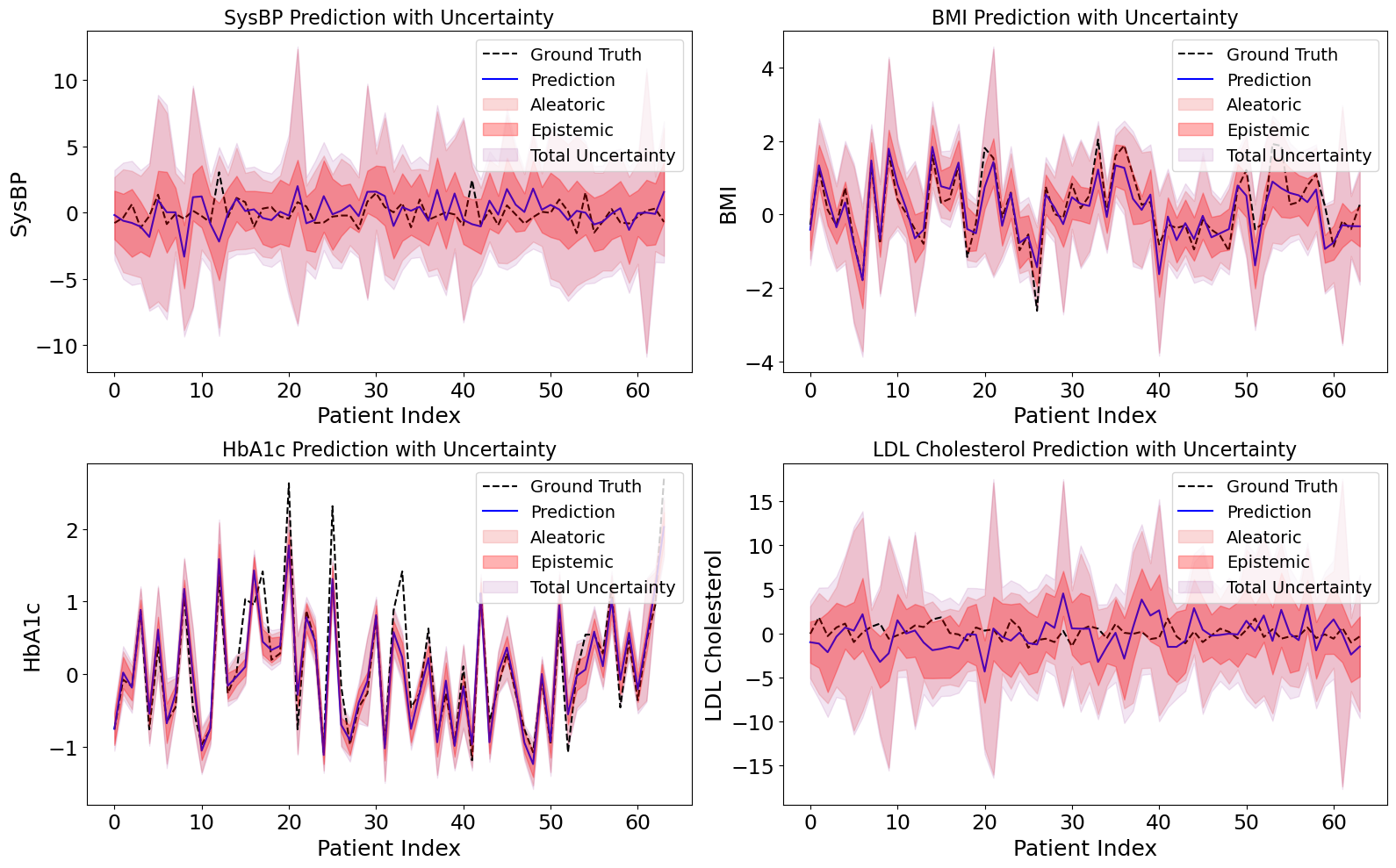}
\caption{Uncertainty Estimation of MBT-CB}
\label{fig:uncertainty_plot}
\end{figure*}

Figure~\ref{fig:uncertainty_plot} presents MBT-CB’s predictions with ground truth and uncertainty bands for four CVD biomarkers. SysBP and LDL-C show narrow intervals overall, with localized spikes in epistemic uncertainty where predictions deviate—indicating model uncertainty in less familiar regions. BMI and HbA1c display broader, aleatoric-dominated bands, suggesting higher intrinsic variability or measurement noise. 

\section{Discussion}

\textbf{MBT-CB outperforms other pre-trained transformer models} enhanced with Bayesian and DeepMTR components. ClinicalBERT's superior performance may be attributed to its richer pre-training on over 1.2 billion clinical tokens and 3M EHRs, compared to smaller corpora used by Bio\_ClinicalBERT (880M), BERT (110M), and MedBERT (57.5M). This broader clinical language exposure improved generalization to our heteroscedastic dataset, aligning with findings in \citep{huang2019clinicalbert}.

\textbf{Ablation results highlight the value of MBT-CB’s components}, particularly Bayesian self-attention and DeepMTR. Bayesian attention enhanced generalization by modeling epistemic uncertainty, improving robustness to sparse or noisy data. DeepMTR captured shared and target-specific patterns across correlated biomarkers, enabling the model to learn biomarker interdependencies and improve predictive accuracy on complex EHR data.

\textbf{Strengths of the MBT-CB Architecture}
MBT-CB integrates pretrained ClinicalBERT, Bayesian self-attention, and a DeepMTR architecture to address challenges in longitudinal EHR-based biomarker prediction. Benefiting from ClinicalBERT’s large(1.2 billion)biomedical and EHR pretraining, the model encodes temporally ordered biomarker prompts using visit-level tokenization, with positional and segment embeddings to track visit order and pandemic phases. 

Bayesian self-attention (12 heads, latent dim 128) models attention weights as Gaussian distributions, with variational inference learning $\mu$ and $\log\sigma$, and KL regularization ($1 \times 10^{-4}$) to capture epistemic uncertainty and reduce overfitting on sparse, irregular data. 

For multi-target prediction, DeepMTR employs a shared fully connected block (512 → 128 → 64) and four output heads for SysBP, BMI, HbA1c, and LDL-C. Demographic, positional, and segment embeddings support personalized, temporal modeling. MBT-CB was trained for 50 epochs (batch size 4, weight decay 0.01) using Hugging Face’s Trainer. 

It consistently outperformed baseline transformers, demonstrating the value of combining pretrained models with Bayesian reasoning and multi-target learning for robust, personalized clinical prediction. These findings underscore the potential of hybrid transformer-based models for advancing personalized prediction in real-world healthcare settings.

\section{Limitations and Future Directions}
 The training data, drawn from two hospitals in 
 Central Massachusetts 
 % [*blinded for review*]
 with 89\% White patients, limits generalizability of our findings. Future work will include multi-site, diverse patients. While Bayesian attention enhances uncertainty estimation, its stochastic nature complicates interpretation. %Attention heatmaps help but may lack actionable insights for clinicians. 
 Integrating SHAP values or counterfactual reasoning can improve explainability.

Future directions include: (1) scaling to larger, more heterogeneous EHR datasets; (2) replacing ClinicalBERT with a custom transformer trained on broader clinical corpora; (3) incorporating SHAP-based attribution to clarify feature contributions to COVID-related biomarker shifts; and (4) benchmarking MBT-CB against deep learning models auto-generated using Network Architecture Search (NAS), and CVD-specific models.

\section{Conclusion }
We presented MBT-CB, a transformer-based model for predicting LDL-C, HbA1c, BMI, and SysBP biomarkers during the COVID-19 pandemic. Leveraging ClinicalBERT,  Bayesian self-attention for uncertainty, multi-target regression and temporal patterns, MBT-CB outperformed all baselines. It produced clear uncertainty bands—narrow with epistemic spikes for SysBP and LDL-C, broader and aleatoric-dominated for BMI and HbA1c—highlighting its ability to distinguish model uncertainty from data noise. This enhances interpretability at the patient level. Future work includes real-time deployment, broader generalization, and SHAP-based explanations.

% \section{Declaration of competing interest}
% The authors declare that they have no known competing financial interests or personal relationships that could have appeared to influence the work reported in this paper.

% \section{Funding sources}
% This research did not receive any specific grant from funding agencies in the public, commercial, or not-for-profit sectors.




% %% For numbered reference style

%   \textit{\LaTeX: a document preparation system},
%   Addison Wesley, Massachusetts,

% \end{thebibliography}

% \newpage
% \appendix
% \section{Example Appendix Section}

% %% If you have bib database file and want bibtex to generate the
% %% bibitems, please use
% %%
% \bibliographystyle{elsarticle-num} 
% \bibliography{references}

\appendix
\section{My Appendix}
\subsection{\textbf{The Vanilla Encoder-only Transformer}}
\label{sec:van_transformer}
\begin{figure*}[htbp]
\centering
\includegraphics[width=0.85\linewidth]{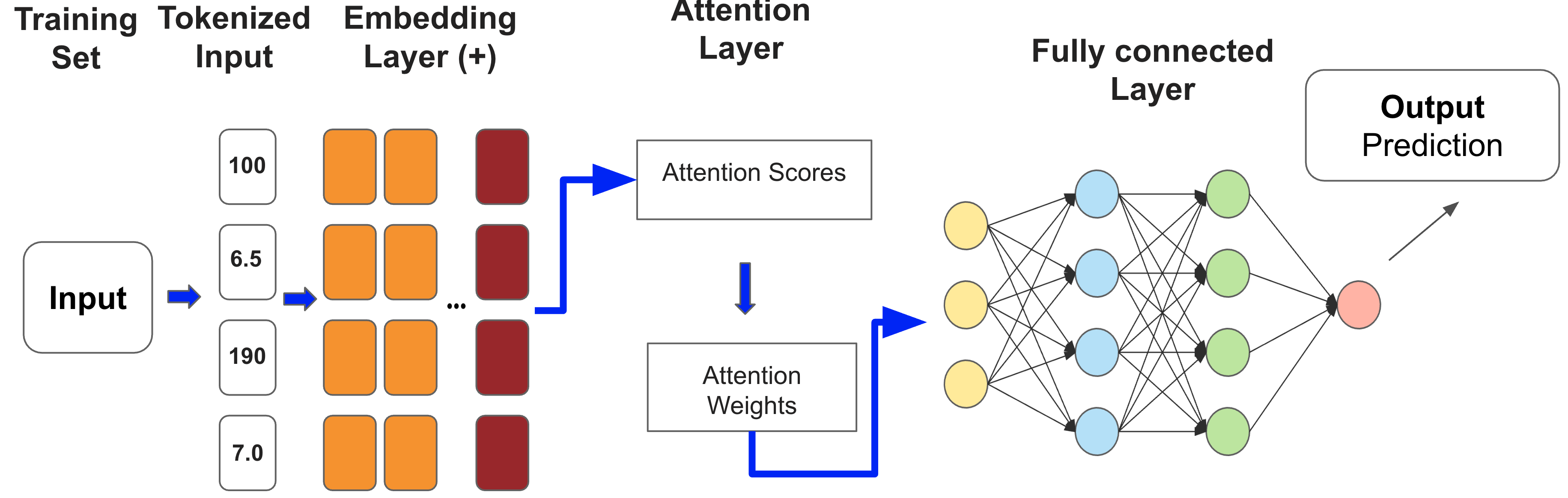}
\caption{A simplified example illustration of an Encoder-only Transformer model}
\label{fig:basic_transf_with_bay_model}
\end{figure*}

Transformers can be either encoder-based, for example the Bi-directional Encoder Representations Transformer (BERT) model or decoder-based, such as Generative Pre-trained Transformer (GPT). Figure~\ref{fig:basic_transf_with_bay_model} illustrates the high-level structure of an encoder-based transformer model. From the image, biomarker sequences for each patient are tokenized and mapped into a high-dimensional embedding space. Each token is then transformed into three vectors (Query, Key, and Value)—via learned linear projections. These vectors are used within the attention layer to determine which parts of the sequence are most relevant to each other. Identified relevancies (attention weights) are eventually multiplied by a value matrix and passed to the fully connected layer for further learning and prediction.

\begin{figure}[htbp]
\centering
\includegraphics[width=0.9\linewidth]{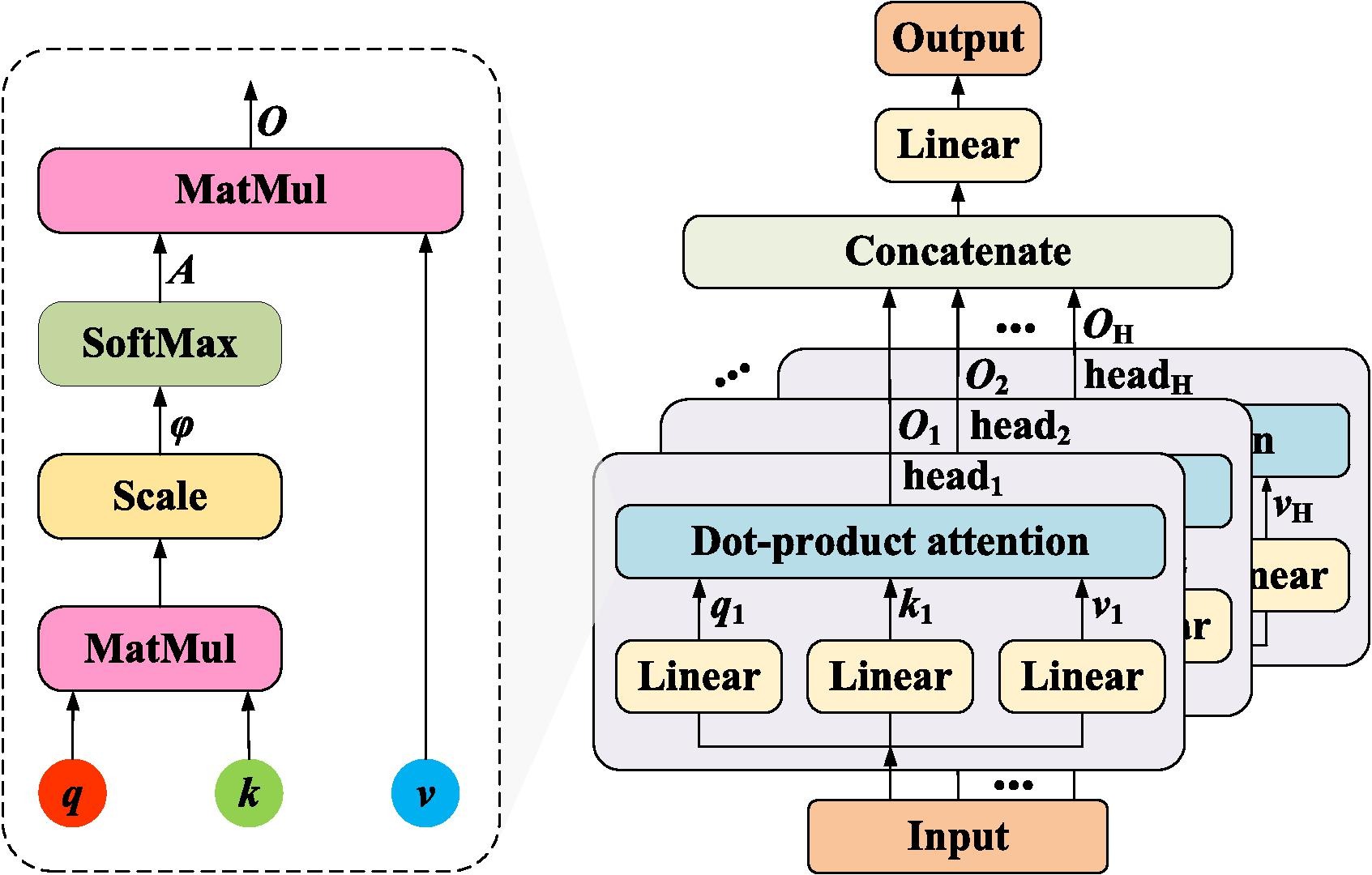}
\caption{Multi-head self-attention mechanism. Image from \citep{xiao2024bayesian}}
\label{fig:transformer_multihead}
\end{figure}

A significant aspect of the Transformer architecture is the multi-head self attention \citep{xiao2024bayesian} which comprises multiple parallel self attention heads, each consisting of linear transformations and dot-product operations (see Figure \ref{fig:transformer_multihead}). Given a set of learned query weights $W^q \in \mathbb{R}^{d_m \times d_k}$, learned key weights $W^k \in \mathbb{R}^{d_m \times d_k}$, learned values weights $W^v \in \mathbb{R}^{d_m \times d_v}$ and input embedding $I \in \mathbb{R}^{m \times d_m}$, the transformer input values (queries, keys, and values) fed into the attention layer are obtained through linear transformations: queries $q = IW^q \in \mathbb{R}^{m \times d_k}$, keys $k = IW^k \in \mathbb{R}^{m \times d_k}$ and values $v = IW^v \in \mathbb{R}^{m \times d_v}$. As a first step, q and k are multiplied via the dot product to get matrix $\varphi$: $\varphi = f_{dot}(q,k) = qk^T / (d_k)^{1/2} \in \mathbb{R}^{m \times m}$.

$(d_k)^{1/2}$ is used to scale matrix values. Next, matrix scores are normalized using the Softmax function on $\varphi$, A = softmax$(\varphi)$ to derive attention weights. These normalized weights are then multiplied by the value matrix to obtain attention results $O = Av \in \mathbb{R}^{m \times d_v}$. Finally, the attention results from all the self attention heads are concatenated to obtain the final attention value $O_{final} = Concat (O_1, O_2, O_3,...,O_H)W^O \in R^{m \times d_m}$. The concatenated attention value matrix from the multi-head self attention are then passed into a linear layer to transform the large matrix into form suitable for input into a NN layer for more learning and prediction. More details is illustrated in Figure \ref{fig:transformer_Var_inf_Att}.
In our BMT model, we extend the basic transformer architecture by introducing Bayesian self-attention where deterministic attention weights are replaced with stochastic distributions. However, to better understand this concept, in the next section we explain bayesian methods in DL.

\begin{figure}[htbp]
  \centering
  
  % First subfigure (on top)
  \begin{subfigure}{\linewidth}
    \centering
    \includegraphics[width= 0.5\linewidth]{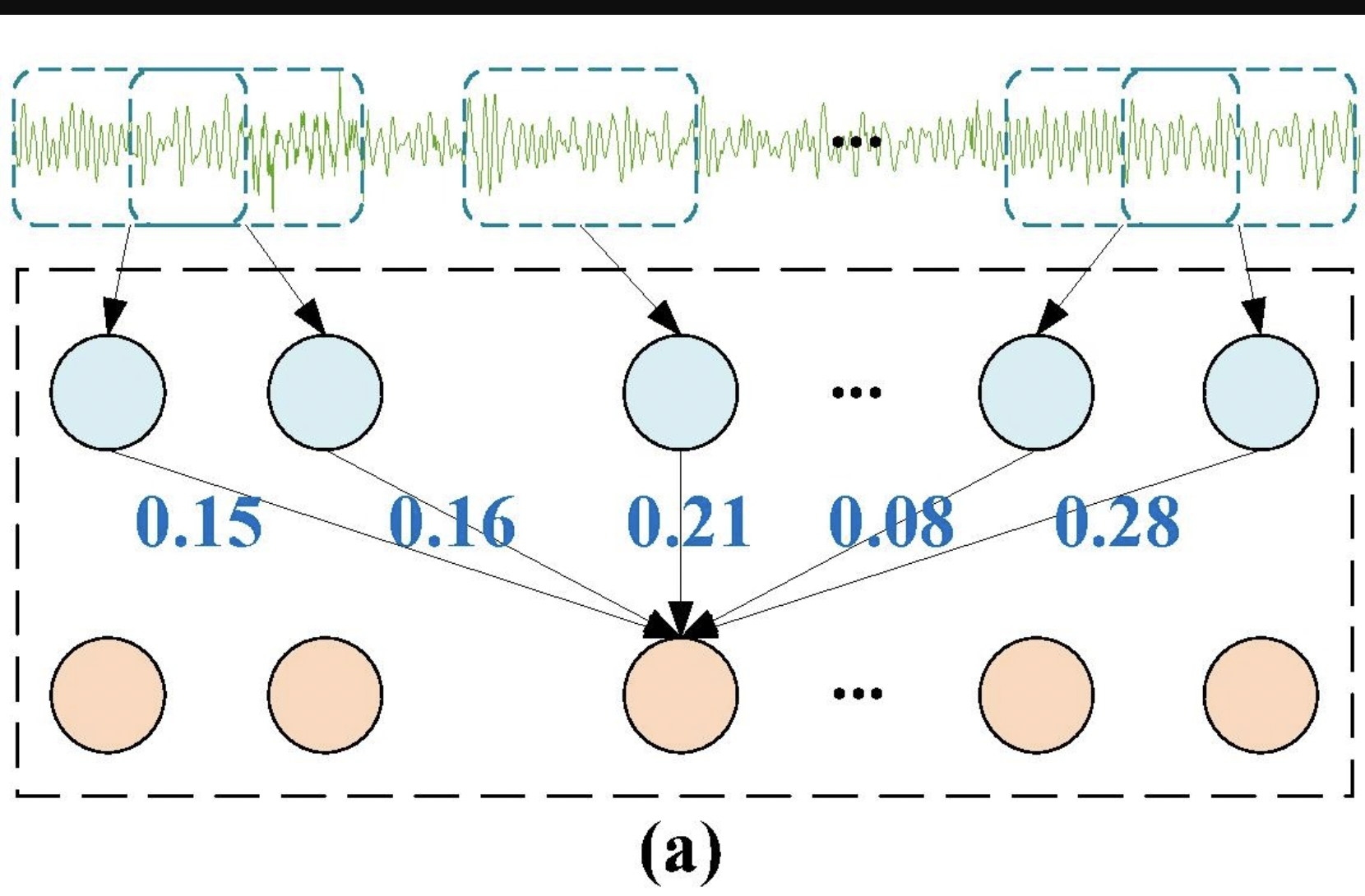}
    \caption{Deterministic attention weights}
    \label{fig:deterministic_attention}
  \end{subfigure}
  
  \vspace{1em}
  
  % Second subfigure (below)
  \begin{subfigure}{\linewidth}
    \centering
    \includegraphics[width= 0.5\linewidth]{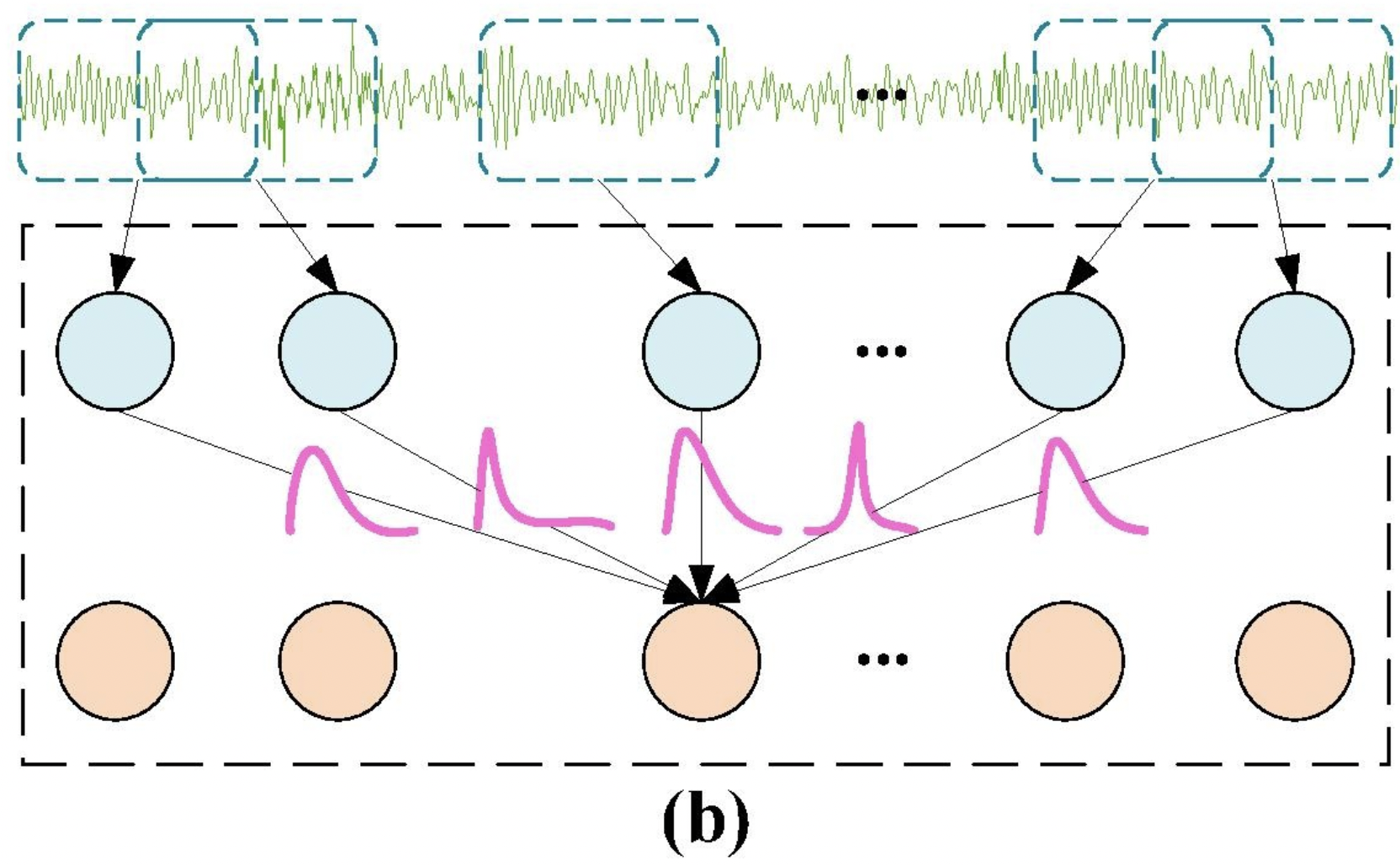}
    \caption{Variational (distributional) attention weights}
    \label{fig:variational_attention}
  \end{subfigure}
  
  \caption{Comparison between deterministic and variational attentions. (a) Deterministic attention; (b) Variational attention. Image adapted from \citep{xiao2024bayesian}.}
  \label{fig:deterministic_and_probabilistic_weights}
\end{figure}

\subsection{\textbf{Bayesian Methods in DL}}
\label{sec:bayesianNN}
Bayesian methods in DL incorporate uncertainty by treating the weights of the network as probability distributions rather than fixed/deterministic values. This probabilistic treatment allows the model to account for uncertainty in both the data and the learned parameters, resulting in more robust and informative predictions. An example of this transformation is as shown in Appendix, Figure \ref{fig:variational_attention}.
In the bayesian method for NN, weights $w$ are initially assumed to follow a distribution called prior distribution $p(w)$, which is then multiplied by the  likelihood $p(D|w)$ and divided by the marginal likelihood $p(D)$. The result gives you a posterior distribution $p(w|D)$, which is the posterior distribution gotten after training on the data (via likelihood), minimizing loss and leveraging the prior knowledge/distribution. This process is governed by Bayes' theorem shown in equation \ref{eq:bayesFormula}:

\begin{equation}
    p(w|D) = \frac{p(D|w) p(w)}{p(D)}
    \label{eq:bayesFormula}
\end{equation}
Here, $p(w)$ is the prior distribution over weights, $p(D|w)$ is the likelihood of the data given the weights, $p(D)$ is the marginal likelihood (also known as evidence), and $p(w|D)$ is the resulting posterior after observing data.
While the posterior distribution $p(w|D)$ captures the complete uncertainty in the model's parameters, it is often intractable to estimate considering the large number of weights that can be present in a NN model, hence various approximation approaches are used including variational inference or Markov Chain Monte Carlo (MCMC) methods.

\subsubsection{\textbf{Variational Inference for Attention Weights}}
\label{sec:var_inference}

In variational inference, we aim to approximate the true posterior distribution \( p(w \mid D) \) by introducing a simpler, tractable distribution \( q(w) \) that is close enough to it. The objective is to minimize the Kullback-Leibler (KL) divergence between the two distributions:

\begin{equation}
    \text{KL}(q(w) \| p(w \mid D)) = \int q(w) \log \left( \frac{q(w)}{p(w \mid D)} \right) dw
\end{equation}

Minimizing this divergence allows us to efficiently estimate the posterior while still capturing meaningful uncertainty. This forms the foundation of \textit{Bayesian NN}, which not only generate predictions but also provide well-calibrated uncertainty estimates—an essential feature in high-stakes applications such as healthcare. In our model, we apply variational inference specifically to the \textit{attention weights}, treating them as latent random variables. The goal is to approximate the true posterior distribution over attention weights \( p(A \mid x, y) \) using a variational distribution \( q_{\theta}(A) \). According to Bayes’ theorem:

\begin{equation}
    p(A \mid x, y) = \frac{p(x, y \mid A) \, p(A)}{p(x, y)}
    \label{eq:bayesFormulaAttention}
\end{equation}

However, since the marginal likelihood \( p(x, y) \) is typically intractable, we instead minimize the KL divergence between the variational approximation and the true posterior:
\begin{align*}
    & KL\left(q_{\phi}(A)\|p(A|x, y)\right) =\int q_{\phi}(A) \log \frac{q_{\phi}(A)}{p(A|x, y)} \, dA \\
    &= \int q_{\phi}(A) \log \frac{q_{\phi}(A)}{p(x, y) p(x, y|A)p(A)} \, dA \\
    &= \log p(x, y) - \underbrace{\int q_{\phi}(A) \log \frac{p(x, y|A)p(A)}{q_{\phi}(A)} \, dA}_{L(x, y)}
\end{align*}

This formulation reveals that minimizing the KL divergence is equivalent to maximizing the \textit{Evidence Lower Bound (ELBO)}, defined as:

\begin{align*}
    \mathcal{L}(x, y) &= \mathbb{E}_{q_{\theta}(A)} \left[ \log p(x, y \mid A) \right] - \text{KL} \left( q_{\theta}(A) \| p(A) \right)
\end{align*}

By maximizing the ELBO during training, we encourage the learned attention weight distribution \( q_{\theta}(A) \) to approximate the true posterior \( p(A \mid x, y) \) as closely as possible, enabling principled uncertainty quantification in our Bayesian Multi-Target Transformer model.

\printcredits

\bibliographystyle{cas-model2-names}

\bibliography{references}


% Author biography without author photo.
% Author biography. Author biography. Author biography.
% Author biography. Author biography. Author biography.
% Author biography. Author biography. Author biography.
% Author biography. Author biography. Author biography.
% Author biography. Author biography. Author biography.
% Author biography. Author biography. Author biography.
% Author biography. Author biography. Author biography.
% Author biography. Author biography. Author biography.
% Author biography. Author biography. Author biography.

% \bio{figs/cas-pic1}
% Author biography with author photo.
% Author biography. Author biography. Author biography.
% Author biography. Author biography. Author biography.
% Author biography. Author biography. Author biography.
% Author biography. Author biography. Author biography.
% Author biography. Author biography. Author biography.
% Author biography. Author biography. Author biography.
% Author biography. Author biography. Author biography.
% Author biography. Author biography. Author biography.
% Author biography. Author biography. Author biography.

% \bio{figs/cas-pic1}
% Author biography with author photo.
% Author biography. Author biography. Author biography.
% Author biography. Author biography. Author biography.
% Author biography. Author biography. Author biography.
% Author biography. Author biography. Author biography.

\end{document}